 \documentclass[final,1p,times]{elsarticle}

\usepackage{amssymb}
\usepackage[latin1]{inputenc}
\usepackage{graphicx}
\usepackage{multirow}
\usepackage{amsmath,amsthm}
\usepackage{subfigure}
\usepackage{amsfonts,amssymb}

 \biboptions{comma,square,sort}

\journal{Fuzzy Sets and Systems}

\graphicspath{{/}}

\begin{document}

\begin{frontmatter}



{\footnotesize\textbf{NOTICE: This is the author's version of a work that was accepted for publication in \emph{Fuzzy Sets and Systems}. Changes resulting from the publishing process, such as peer review, editing, corrections, structural formatting, and other quality control mechanisms may not be reflected in this document. Changes may have been made to this work since it was submitted for publication. A definitive version was subsequently published in \emph{Fuzzy Sets and Systems}, 234, 1 (January 2014) DOI:10.1016/j.fss.2013.02.007.} }

\title{A Fuzzy Syllogistic Reasoning Schema for Generalized Quantifiers}

\author{M. Pereira-Fariña\corref{*}}
\ead{martin.pereira@usc.es}
\cortext[*]{Corresponding author. Tel.: +0034 881812726; fax: +0034 881 813602}

\author{Juan C.~Vidal}
\ead{juan.vidal@usc.es}

\author{F. D\'iaz-Hermida}
\ead{felix.diaz@usc.es}

\author{A. Bugar\'in}
\ead{alberto.bugarin.diz@usc.es}

\address{Centro de Investigaci\'on en Tecnolox\'ias da Informaci\'on (CITIUS), University of Santiago de Compostela, Campus Vida, E-15782, Santiago de Compostela, Spain}

\begin{abstract}
In this paper, a new approximate syllogistic reasoning schema is described that expands some of the approaches expounded in the literature into two ways: (i) a number of different types of quantifiers (logical, absolute, proportional, comparative and exception) taken from \textit{Theory of Generalized Quantifiers} and similarity quantifiers, taken from statistics, are considered and (ii) any number of premises can be taken into account within the reasoning process. Furthermore, a systematic reasoning procedure to solve the syllogism is also proposed, interpreting it as an equivalent mathematical optimization problem, where the premises constitute the constraints of the searching space for the quantifier in the conclusion.
\end{abstract}


\end{frontmatter}

\section{Introduction}
Human beings usually manage statements or propositions that involve quantities that are more or less well defined. Quantifiers, such as \textit{all, few, 25, around 25\dots}, are the linguistic particles frequently used to express them. Quantified statements (sentences involving quantifiers) are used both for describing particular aspects of reality (i.e., \textit{Most students are young}) and also for making inferences; that is, obtaining new information from a given set of premises. This type of reasoning is known as \textit{syllogism}. Although syllogisms were superseded by propositional logic~\cite{Kneale1968} in the 19th century, it is still matter of research. In this paper, we propose an approach to syllogism involving fuzzy generalized quantifiers that allows the reasoning to be performed with no limits in the number of premises.\par
 
\begin{table}[htb!]
	\begin{center}
	\caption{An example of syllogism.}
	\label{tab:FuzzySyllogism}
	    \vspace{2mm}
		\begin{tabular}{cl} 
			$PR1$ & Eight students are Portuguese\\ 
			$PR2$	& Ten students are young\\ \hline
			$C$ 	& Eight or less students are Portuguese and young
		\end{tabular}
	\end{center}
\end{table}

Table~\ref{tab:FuzzySyllogism} shows an example of an usual syllogism, made up of two premises and one conclusion. In this case the premises are the quantified statements denoted as $PR1$ (first premise) and $PR2$ (second premise) and the conclusion is the quantified statement denoted as $C$ (inferred from the premises). Each one of the quantified statements in the example is made up of two main elements: a quantifier (\textit{eight, ten, eighteen or less}) and terms (\textit{students, Portuguese, young, Portuguese and young}), usually interpreted as sets, that describe properties of the elements in a referential universe. The most usual quantified statements are the binary ones; that is, statements involving a single quantifier and two terms. The subject in the sentence is the ``restriction'' of the quantifier and the predicate its ``scope''. For instance, \textit{students} is the restriction and \textit{Portuguese} the scope of quantifier \textit{eight} in $PR1$ premise of Table~\ref{tab:FuzzySyllogism}. The purpose of inference within this context is to calculate consistent values for the quantifier in the conclusion starting from the premises. These values are strongly dependent on the actual distribution of the elements in the referential among the terms\footnote{In~\cite{Kumova2010}, an extension of Aristotelian syllogistics considering the problem of the distribution is proposed.}. In the example, the consistent values for the quantifier in the conclusion range from 0, for the case in which none of the eight Portuguese students is young, to 8, for the case in which all of the Portuguese students are also young.\par 



Syllogistic reasoning is an interesting field from different points of view, that include theoretical and applied ones. From a theoretical perspective, as a logic, it is a kind of reasoning that should be analyzed and understood in order to increase our knowledge about how human beings perform common-sense reasoning in daily life. From an applied perspective, it is an interesting tool for the fields of decision making or database systems~\cite{Liu1998}.
Furthermore, our approach to syllogistic reasoning can be contextualized in the Computing with Words paradigm, because we try to manage a form of common-sense reasoning preserving the natural language surface.\par

On the other hand, it is relevant to note that, from the point of view of reasoning, we usually focus on situations where quantified statements are assumed to be true and we are only concerned about what types of information can be inferred from them \cite{Yager1986}.  A different problem is the evaluation of quantified statements where the information about the fulfillment of the involved terms in a given universe is available. Models for this problem \cite{Diaz2003,Glockner2006} do not consider the topic of syllogistic reasoning.\par

The first systematic approach to syllogistic reasoning was developed in~\cite{Aristotle1949}. Nevertheless, this framework only deals with arguments composed by two premises and one conclusion and only handles the four classical logic quantifiers (\textit{all, none, some, not all}). Therefore, expressiveness of this model is limited to simple statements that are far from the common uses of language.\par

Most of approaches found in the literature try to extend syllogistic reasoning in two parallel ways:

\begin{itemize}
	\item by adding new quantifiers (like \textit{most, few, many}, and so on) to the classical ones, considering crisp~\cite{Peterson2000} and fuzzy~\cite{Zadeh1985,Dubois1988,Dubois1990,Dubois1993,Novak2008,Schwartz1997} definitions. In these approaches only absolute/relative quantifiers involving two terms are handled, and many of the classical syllogistic reasoning patterns are not even considered~\cite{Pereira2012}
	
	\item by considering arguments composed by $N$ statements and $N$ terms but limited to the four logic quantifiers~\cite{Sommers1982}, therefore missing much of the actual expressive necessities of daily language and reasoning
\end{itemize}

These two ways have been approached in a mutually exclusive way, probably due to the fact that the first approaches come from the field of fuzzy modeling and the second ones from the field of linguistics and natural logic. To the best of our knowledge, no combination of both points of view has been made so far for providing a meaningful extension of the syllogistic patterns that at the same time involve relevant quantifiers (as those described in the linguistics field) and more than two terms in the statements. Therefore a general approach to syllogistic reasoning remains an incomplete task, since these approaches deal with syllogistics only from partial perspectives. Within this context, the main aim of this paper is to present a general formulation of syllogistic reasoning and its resolution that can be useful for all the fields aforementioned.
Our formulation of syllogism is capable of managing different types of quantifiers within the \textit{Theory of Generalized Quantifiers} (TGQ) (going much further than the usual absolute/relative quantifiers) and $N>2$ premises in the reasoning scheme. Thus, our model can deal with more complex arguments and more wide fragments of natural language. Notwithstanding, the combination of absolute/proportional quantifiers in the same syllogism remains an open question and it is not dealt with in this paper.\par


The structure of the paper is as follows: in section~\ref{sc:LiteratureReview} we present the most relevant approaches to fuzzy syllogistic reasoning; in section~\ref{sc:GeneralSyllogisticReasnongPattern} the general form of our proposal of syllogistic schema and its resolution are detailed; in section~\ref{sc:ExamplesSyllogisms} some illustrative examples are shown. Finally, in section~\ref{sc:Conclusions}, we summarize the conclusions of the paper.

\section{Literature review}
\label{sc:LiteratureReview}

In~\cite{Spies1989} a relevant distinction in the analysis framework of the fuzzy syllogism is pointed out. There are two possible interpretations considering quantified statements with the basic structure ``$Q$ $A$ are $B$'': (i) \textit{relative-frequency} rules, where the terms $A$ and $B$ denote sets over a given universe and the quantifier $Q$ denotes the relationship between both sets (e.g.; \textit{A few doctors are detective consulting}); (ii) \textit{conditional} rules, where $A$ is the antecedent; $B$ is the consequent and $Q$ denotes the power of the link ($\rightarrow$) between the antecedent and the consequent (e.g.; \textit{If J. Watson is a doctor, then it is unlikely that he is a detective consulting}). It is worth noting that the typical linguistic structure of this kind of rules is ``If $A$ then $B$ $[a,b]$'' where the interval $[a,b]$ denotes the reliability or the power of the link.\par

Both interpretations of the rules behave differently when they are applied to syllogistic reasoning (see Table~\ref{tab:SpiesSyllogisms}). Interpretation (i) uses Zadeh's approach, and can be exactly used for syllogistic reasoning since the statements maintain the typical structure used in natural language. Interpretation (ii), based on support logic, is closer to the \textit{fuzzy modus ponens} than the syllogistic reasoning.\par

\begin{table}[tb!]
	\centering
	\caption{Spies' syllogisms~\cite{Spies1989}.}
	\label{tab:SpiesSyllogisms}
	    \vspace{2mm}
	  \subtable{
		\small
		\begin{tabular}{c l}
		\hline
			\multicolumn{2}{c}{\textbf{Relative-frequency example}}\\\hline\hline
			$PR1$ & $Q_{1}$ dogs bark cats\\ 
			$PR2$	& $Q_{2}$ pets are dogs										\\\hline
			$C$ 	& $Q$ pets bark cats	
		\end{tabular}
		}
		\subtable{
		\small
		\begin{tabular}{c l}
		\hline
		\multicolumn{2}{c}{\textbf{Conditional  example}}\\\hline\hline
		$PR1$ & If Jim is not at home, he is in his office $[a,b]$\\
		$PR2$	& If Jim is working, he is not at home $[c,d]$\\\hline
		$C$ 	& If Jim is working, he is in his office $[f(a,b),f(c,d)]$
		\end{tabular}
		}
\end{table}

On the other hand, syllogisms are classified into two classes~\cite{Spies1989}:
\begin{enumerate}  
	\item Property inheritance (asymmetric syllogism): A term-set $X$ and a term-set $Z$ are linked via concatenation of $X$ with a term-set $Y$ and $Y$ with $Z$:\par
	\vspace{2mm}
	\begin{center}
		\begin{tabular}{rl}
			$PR1:$& $Y$ in relation to $Z$\\
			$PR2:$& $X$ in relation to $Y$\\
			$C:$  & $X$ in relation to $Z$
		\end{tabular}
	\end{center}
	The position of $Y$ can change in the premises; depending on its position, the four Aristotelian figures appear~\cite{Kneale1968}.
	\vspace{2mm}
	\item Combination of evidence (symmetric syllogism): The ties between $X$ and $Z$ and between $Y$ and $Z$ are calculated separately and both are joined in the conclusion by a logic operator (conjunction/disjunction):\par
	\vspace{2mm}
	\begin{center}
	\begin{tabular}{rl}
			$PR1:$& $Y$ in relation to $Z$\\
			$PR2:$& $X$ in relation to $Z$\\
			$C:$  & $X \& Y$ in relation to $Z$
		\end{tabular}
	\end{center}
	\end{enumerate}

In~\cite{Schwartz1996,Schwartz1997} a different definition of syllogistic reasoning is proposed. Here, a ``qualified syllogism'' is defined as a ``classical Aristotelian syllogism that has been `qualified' through the use of fuzzy quantifiers, likelihood modifiers, and usuality modifiers''~\cite{Schwartz1996}. Nevertheless, the reasoning scheme that the author proposes (see Table~\ref{tab:SchwartzSyllogism}) does not satisfy the Aristotelian definition since neither $PR2$ nor $C$ have the appropriate form.\par

\begin{table}[tb!]
	\begin{center}
	\caption{Schwartz's syllogism~\cite{Schwartz1997}.}
	\label{tab:SchwartzSyllogism}
	    \vspace{2mm}
		\begin{tabular}{cl} 
			$PR1$ & Most birds can fly\\ 
			$PR2$	& Tweety is a bird\\ \hline
			$C$ 	& It is likely that Tweety can fly
		\end{tabular}
	\end{center}
\end{table}

Another approach to syllogistic reasoning in the fuzzy field was developed in~\cite{Novak2008,Murinova2012}, where a fuzzy-logic formalization of intermediate quantifiers \cite{Peterson2000} and reasoning with them is presented. Within this fuzzy type theory -the higher order fuzzy logic- syntactic proofs on the validity of 105 Aristotelian syllogisms involving intermediate quantifiers are provided. \par

Other relevant approach to syllogistic reasoning was proposed in~\cite{Zadeh1983,Zadeh1985} and in~\cite{Yager1986}. Zadeh defines in~\cite{Zadeh1985} the fuzzy syllogism as ``\textit{...an inference scheme in which the major premise, the minor premise and the conclusion are propositions containing fuzzy quantifiers}''. A typical example of Zadeh's fuzzy syllogism is shown in Table~\ref{ex:Zadehfuzzysyllogism}. The quantifier of the conclusion, $Most \otimes Most$ is calculated from the quantifiers in the premises by applying the \textit{Quantifier Extension Principle} (QEP)~\cite{Zadeh1983}, in this case using the fuzzy arithmetic product.\par 
	\begin{table}[htb]%
	\centering
	\caption{\label{ex:Zadehfuzzysyllogism} Zadeh's syllogism~\cite{Zadeh1985}.}
	\vspace{2mm}
	      \begin{tabular}{cl}%
		  $PR1$ & Most students are young\\
		  $PR2$ & Most young students are single\\\hline
		  $C$ 	& $Most \otimes Most$ students are young and single
	      \end{tabular}	  
	\end{table}

Zadeh's approach is characterized by the interpretation of fuzzy quantifiers as fuzzy numbers~\cite{Zadeh1983}. It is worth noting that he only deals with absolute/proportional quantifiers, being absolute quantifiers (\textit{around 25, none,\dots}) identified with absolute fuzzy numbers and proportional quantifiers (\textit{a few, many, most,\dots}) with proportional fuzzy numbers. Other types of quantifiers inspired in the linguistics area~\cite{Diaz2003}, such as comparative (\textit{there are about 3 more tall people than blond people}) or exception (\textit{all except 3 students are tall}) are therefore excluded. On the other hand, Zadeh defines a number of syllogistic patterns~\cite{Zadeh1985} where fuzzy arithmetics is applied to calculate the quantifier of the conclusion. As we said before, all Zadeh's patterns are based on the QEP. This principle is a special case of the most general entailment principle~\cite{Zadeh1987} and claims that if there exists a functional relation $f$ between the conclusion and the premises of a syllogism, then a function $\phi$ is defined as an extension of $f$ by the extension principle. Nevertheless, as pointed out in~\cite{Liu1998}, the QEP shows some disadvantages. Firstly, different results can be obtained depending on how the cardinality equations are written; in consequence, ambiguity arises in the result of a syllogism\footnote{For instance, in the example of Table~\ref{ex:Zadehfuzzysyllogism}, if $most=\frac{3}{4}$, the quantifier of $C$ is $\left(\frac{3}{4} \right)^{2}$. This functional relationship can also be written as $\frac{most^{3}}{most}$, but in the case of fuzzy arithmetics with fuzzy numbers the equivalence between $most^{2}$ and $\frac{most^{3}}{most}$ is not fulfilled.}. Moreover, inference schemes are not reverse either with respect to addition or multiplication. Finally, there is an incompleteness of fuzzy quantifiers and fuzzy numbers, and, therefore, a lot of chances of getting a discontinuous result~\cite{Liu1998}. Finally, in~\cite{Pereira2010,Pereira2012b} it is shown how some of the classical syllogisms are not considered and cannot be adequately managed by Zadeh's approach.\par
	
Yager's approach~\cite{Yager1986} is an extension of Zadeh's one.  In~\cite{Yager1986} some additional patterns maintaining a limited number of premises are included. The proposal has the same limitations of Zadeh's one.\par

To the aim of this paper, the most interesting approach is proposed by D. Dubois et al.~\cite{Dubois1988,Dubois1990,Dubois1993} where quantifiers are interpreted as crisp closed intervals (\textit{more than a half}$=[0.5,1]$, \textit{around three}$=[2,4]$). This approach is mostly focused to proportional quantifiers (such as \textit{most, many, some, between 25\% and 34\%,\dots}) and so a quantifier $Q$ is modelled as $Q:=[\underline{q}, \overline{q}]$, where $\underline{q}, \overline{q} \in \left[0, 1\right]$. Other types of quantifiers are considered by adapting the interval definition to each case: precise quantifiers (those whose values are precisely known and have precise bounds) like $10\%, 30\%,\ldots$ and represented as  a particular case of the previous one (taking $\underline{q} = \overline{q}$); and fuzzy quantifiers (those whose bounds are ill-defined and have imprecise, fuzzy bounds) like \textit{most, few,\dots} represented using fuzzy sets. Regarding the reasoning procedure, three patterns~\cite{Dubois1990} are proposed. Table~\ref{ex:DPfuzzysyllogism} shows an example of Pattern III.

\begin{table}[tbh]%
  \centering
  	\caption{\label{ex:DPfuzzysyllogism} Interval fuzzy syllogism.}
	  \begin{tabular}{cl}%
			$PR1$ & $[5\%, 10\%]$ people that have children are single\\
	  	$PR2$ & $[15\%, 20\%]$ people that have children are young\\\hline
	  	$C$ & $[0\%, 10\%]$ people that have children are young and single
		\end{tabular}	 
\end{table}

In this approach, the reasoning process is transformed into a calculation procedure consisting of the minimization and maximization of the quantifier in the conclusion. This quantifier can be modeled as an interval or a trapezoidal function and is calculated taking the quantifiers of the premises as restrictions. The main aim is to obtain the most favourable and the most unfavourable proportion among the terms of the conclusion according to the proportions expressed in the premises. Using the results described in~\cite{Dubois1988,Dubois1990}, a first step for the development of a fuzzy linguistic syllogism is developed in~\cite{Dubois1993}, where the syllogistic patterns are totally expressed using linguistic terms. This approach avoids fuzzy arithmetics problems of Zadeh's approach but is still limited to a few reasoning patterns and many of the classical syllogistic patterns are not considered~\cite{Pereira2010,Pereira2012b}. Nevertheless, the interpretation of the quantifiers as intervals and the transformation of the reasoning problem into a calculation procedure establishes the basis for our proposal of a general approach to syllogistic reasoning that can manage arguments with $N$ sentences and terms, the two types of syllogisms (property inheritance an combination of evidence) and involving new types of quantifiers inspired in the linguistics area \cite{Diaz2003}, such as the aforementioned comparative or exception quantifiers.\par


\section{A general syllogistic reasoning pattern}
\label{sc:GeneralSyllogisticReasnongPattern}

\subsection{Formulation of quantified statements with generalized quantifiers}
\label{ssc:FormulationQuantifiedStatements}
The TGQ~\cite{Barwise1981} analyzes quantification in natural language combining the logic and the linguistics perspectives. Its key concept is the \textit{generalized quantifier}, understood as a second order predicate that establishes a relationship between two classical sets. As an example, we describe in detail the case of proportional, absolute and exception quantifiers.\par

Considering the proportional interpretation of quantifier \textit{all}, and denoting $E$ as the referential universe and $\mathcal{P}(E)$ as the power set of $E$, the evaluation of the quantified statement ``\emph{All $Y_{1}$ are $Y_{2}$}'' for $Y_{1}, Y_{2} \in \mathcal{P}(E)$ can be modeled as:
\begin{eqnarray*}
all: \mathcal{P}(E) \times \mathcal{P}(E)  & \rightarrow & \left\{0,1\right\}\\
		(Y_{1},Y_{2})					 & \rightarrow & All(Y_{1},Y_{2})= \left\{ 
										  										\begin{aligned}
							          										& 0: if &Y_{1} \not\subseteq Y_{2}\\
										   											& 1: if &Y_{1} \subseteq Y_{2}
							        										\end{aligned} \right.
\label{eq:QuantifierAll}
\end{eqnarray*}

For absolute quantifiers, evaluation of a statement like ``\textit{Between 3 and 6 $Y_{1}$ are $Y_{2}$}'' can be modelled as:
\begin{eqnarray*}
\text{\textit{Between 3 and 6}}: \mathcal{P}(E) \times \mathcal{P}(E)  & \rightarrow & \left\{0,1\right\}\\
																		(Y_{1},Y_{2})& \rightarrow &\text{\textit{Between 3 and 6}}(Y_{1},Y_{2})= \left\{ 
																																			  \begin{aligned}
																																          & 0: if &|Y_{1} \cap Y_{2}| \notin [3,6]\\
																																			   	& 1: if &|Y_{1} \cap Y_{2}| \in [3,6]
																																        \end{aligned} \right.
\end{eqnarray*}

For quantifiers of exception, let us consider the sentence ``\textit{All but 3 $Y_{1}$ are $Y_{2}$}''; for $Y_{1}, Y_{2} \in \mathcal{P}(E)$. Its evaluation can be modelled as:
\begin{eqnarray*}
\text{\textit{All but 3}}: \mathcal{P}(E) \times \mathcal{P}(E) &\rightarrow& \left\{0,1\right\}\\
																	(Y_{1},Y_{2}) &\rightarrow& \text{\textit{All but 3}}(Y_{1},Y_{2})= \left\{ 
																															  \begin{aligned}
																												          & 0: if &|Y_{1} \cap \overline{Y_{2}}| \neq 3\\
																															   	& 1: if &|Y_{1} \cap \overline{Y_{2}}| = 3
																												        \end{aligned} \right.
\end{eqnarray*}

Let $P=\left\{P_{s},\:s=1\dots,S\right\}$ be the set of relevant properties defined in $E$. For instance, $P_{1}$ can denote the property ``to be a student'' and $P_{2}$ ``to be tall''. Let $L_{1}$ and $L_{2}$ be any boolean combination of the properties in $P$. So, the typical statement involved in a syllogism has the following general structure:

\begin{equation}
	Q \textnormal{  } L_{1} \textnormal{ are } L_{2}
\end{equation}

\noindent
where $Q$ denotes a linguistic quantifier, $L_{1}$  is the subject-term or restriction and $L_{2}$ the predicate-term or scope. In this paper, we consider the following types of quantifiers:

\begin{itemize}
	\item \textbf{Logical quantifiers ($Q_{LQ}$)}: The same classical quantifiers managed by Aristotle (\textit{all, none, some, not all}).
	\item \textbf{Absolute Binary quantifiers ($Q_{AB}$)}:  Natural numbers ($N$) with or without some type of modifier (\textit{around 5, more than 25,\dots}) following the general pattern ``$Q_{AB}$ $Y_{1}$ are $Y_{2}$'' (e.g., ``\textit{around 5 students are tall}'').
	\item \textbf{Proportional Binary quantifiers ($Q_{PB}$)}: Linguistic terms ($LT$) (\textit{most, almost all, few,\dots}) following the general pattern ``$Q_{PB}$ $Y_{1}$ are $Y_{2}$'' (e.g. ``\textit{few students are tall}'',\dots).
	\item \textbf{Binary Quantifiers of Exception ($Q_{EB}$)}: Proportional binary quantifiers with a natural number ($N*$) (\textit{all but 3, all but 3 or 4,\dots}) following the general pattern ``$Q_{EB}$ $Y_{1}$ are $Y_{2}$'' (e.g., ``\textit{all but 5 students are tall}'').
	\item \textbf{Absolute Comparative Binary Quantifiers ($Q_{CB-ABS}$)}: Natural numbers with modifiers of type \textit{more, less,} (\textit{3 more\dots than, 4 less\dots than,\dots}) following the general pattern ``There is/are $Q_{CB-ABS}$ $Y_{1}$ than $Y_{2}$'' (e.g. ``\textit{there are 3 more boys than girls}'').
	\item \textbf{Proportional Comparative Binary Quantifiers ($Q_{CB-PROP}$)}: Rational multiple or partitive numbers ($Q*$) \textit{double, half,\dots} following the general pattern ``There is/are $Q_{CB-PROP}$ $Y_{1}$ than $Y_{2}$'' (e.g., ``\textit{there are double boys than girls}'').
  \item \textbf{Similarity Quantifiers ($Q_{S}$)}:	Linguistic expressions that denotes similarity ($S$) between two given sets \textit{very similar, few similar,\dots} following the general pattern ``$A$ and $A'$ are \textit{very/few/\dots} similar''; where $A$ denotes one of the sets of the comparison and $A'$ denotes a set similar to $A$ (e.g. ``\textit{The audience of opera and ballet are very similar}''). 
\end{itemize}

Table~\ref{tab:FamiliesGeneralizedQuantifiers} shows the definition of these quantifiers according to the TGQ. 

\begin{table}[htb]
	\begin{center}
		\caption{Definition of generalized quantifiers.}
		\vspace{2mm}
		\label{tab:FamiliesGeneralizedQuantifiers}
			\begin{tabular}{|rl|} \hline
			\multicolumn{2}{|c|}{\textbf{Logical quantifiers}}\\\hline			
			$Q_{LQ-all} (Y_{1}Y_{2})=$  & $\left\{\begin{array}{ll}
																		0:&Y_{1} \not\subseteq Y_{2}\\
																		1:&Y_{1} \subseteq Y_{2} \end{array} \right.$\\\hline
			$Q_{LQ-no} (Y_{1},Y_{2})=$  & $\left\{\begin{array}{ll}
																		0:&Y_{1} \cap Y_{2} \neq \emptyset \\
																		1:&Y_{1} \cap Y_{2}= \emptyset \end{array} \right.$\\\hline
			$Q_{LQ-some} (Y_{1},Y_{2})=$ & $\left\{\begin{array}{ll} 
																		0:&Y_{1} \cap Y_{2}=\emptyset \\
																		1:&Y_{1} \cap Y_{2}\neq \emptyset \end{array} \right.$\\\hline													
			$Q_{LQ-not-all} (Y_{1},Y_{2})=$ 	& $\left\{\begin{array}{ll}
																		0:&Y_{1} \subseteq Y_{2}\\
																		1:&Y_{1} \not\subseteq Y_{2} \end{array} \right.$\\\hline
			\multicolumn{2}{|c|}{\textbf{Absolute binary quantifiers}}\\\hline
			$Q_{AB}(Y_{1},Y_{2})=$ & $\left\{\begin{array}{ll}
																		0:& |Y_{1} \cap Y_{2}| \neq N \\
																		1:& |Y_{1} \cap Y_{2}| = N \end{array} \right.$\\\hline
			\multicolumn{2}{|c|}{\textbf{Proportional binary quantifiers}}\\\hline
			$Q_{PB}(Y_{1},Y_{2})=$ & $\left\{\begin{array}{ll}
																0:& \frac{|Y_{1} \cap Y_{2}|}{|Y_{1}|} \notin LT\\
																1:& \frac{|Y_{1} \cap Y_{2}|}{|Y_{1}|} \in LT \\
																1:& |Y_{1}|=0 \end{array} \right.$\\\hline
			\multicolumn{2}{|c|}{\textbf{Exception binary quantifiers}}\\\hline
			$Q_{EB}(Y_{1},Y_{2})=$ & $\left\{\begin{array}{ll}
																		0:&|Y_{1} \cap \overline{Y_{2}}| \notin N*\\
																		1:&|Y_{1} \cap \overline{Y_{2}}| \in  N* \end{array} \right.$\\\hline
			\multicolumn{2}{|c|}{\textbf{Comparative binary quantifiers}}\\\hline
			$Q_{CB-ABS}(Y_{1},Y_{2})=$ & $\left\{\begin{array}{ll}
																				0:& |Y_{1}| - |Y_{2}| \notin N*\\
																				1:& |Y_{1}| - |Y_{2}| \in N \end{array} \right.$\\\hline
			$Q_{CB-PROP}(Y_{1},Y_{2})=$ & $\left\{\begin{array}{ll}
																				0:& \frac{|Y_{1}|}{|Y_{2}|} \notin Q*\\
																				1:& \frac{|Y_{1}|}{|Y_{2}|} \in Q* \end{array} \right.$\\\hline
			\multicolumn{2}{|c|}{\textbf{Similarity quantifiers}}\\\hline
			$Q_S(Y_{1},Y_{2})=$ &$ \left\{\begin{array}{ll}
											0:& \frac{|Y_{1}\cap Y_{2}|}{|Y_{1}\cup Y_{2}|} < S, Y_{1}	\cup Y_{2} \neq \emptyset, S\notin Q*\\
											1:& \frac{|Y_{1}\cap Y_{2}|}{|Y_{1}\cup Y_{2}|} \geq S, Y_{1} \cup Y_{2} \neq \emptyset, S\in Q*\\															1:& Y_{1} \cup Y_{2} = \emptyset \end{array} \right.$\\\hline
		\end{tabular}
	\end{center}
\end{table}


\subsection{General inference schema for syllogistic reasoning}
\label{sc:GeneralInferenceSchema}
Reasoning consists of making explicit or extracting the implicit information contained in the premises. The main characteristic of deductive reasoning is to infer new information, a conclusion, from the set of statements that constitutes the premises of the argument. Syllogism, a type of deductive inference, is based on the relationship among sets and their cardinalities. Because of this, the reasoning process is directly linked with the classical distribution problem \cite{Peterson1995}, i.e., how the elements of the referential fulfill the properties or terms in the statements. Thus, from this point of view, each one of the premises of the syllogism is a restriction that delimits the distribution and the conclusion is another constraint that must be compatible with the distribution described in the premises.\par

Assuming this interpretation and following~\cite{Dubois1990}, we propose a transformation of the syllogistic reasoning problem into an equivalent optimization problem that consists of calculating the extreme quantifiers of the conclusion taking as restrictions the quantifiers of the premises. In~\cite{Dubois1990}, this idea is pointed out exclusively considering crisp sets, proportional quantifiers and only some reasoning patterns. However, a more general pattern is needed to support arguments constituted by $N$ premises, $PR_{n}, n=1,\dots, N$ and a conclusion, $C$, following the pattern in expression~(\ref{eq:GeneralPatternSyllogism}):

\begin{equation}
	\begin{tabular}{cl}
		$PR1:$ & $Q_{1} $ $ L_{1,1} $ are $ L_{1,2}$\\
		$PR2:$ & $Q_{2} $ $ L_{2,1} $ are $ L_{2,2}$\\
		\dots\\
		$PRN:$ & $Q_{N} $ $ L_{N,1} $ are $ L_{N,2}$\\ \hline
		$C:$ 	 & $Q_{C} $ $ L_{C,1} $ are $ L_{C,2}$\\
	\end{tabular}
\label{eq:GeneralPatternSyllogism}
\end{equation}

\noindent
where $Q_{n}$, $n=1,$\dots $,N$ are the linguistic quantifiers in the $N$ premises; $L_{n,j},\:n=1,\dots,N, \:j=1,\:2$ denote an arbitrary boolean combination between the properties considered in the syllogism, $Q_{C}$ stands for the quantifier of the conclusion and $L_{C,1}$ and $L_{C,2}$ stand for the subject-term and the predicate-term in the conclusion. In addition, it is worth to mention that both the asymmetric and symmetric syllogisms are consistent with this general definition.


In order to support the general inference schema described in (\ref{eq:GeneralPatternSyllogism}), we distinguish three steps: (i) dividing the universe of discourse into disjoint sets; (ii) defining sentences as systems of inequations; (iii) selecting the optimization method that should be applied in each case in order to resolve the reasoning process.

\subsubsection{Division of the universe in disjoint sets}
\label{ssc:DivisionDisjointSets}
The referential universe $E$ is partitioned into a new set of disjoint sets $PD=\left\{P'_{1},\dots,P'_{K}\right\}$ with $K=2^{S}$ containing the elements in $E$ that fulfill or not the $S$ properties in the following way:\par
\begin{equation}
 	  \begin{aligned}
        & P'_{1}= \overline{P_{1}} \cap \overline{P_{2}} \cap \dots	\overline{P_{S-1}} \cap \overline{P_{S}}\\
		   	& P'_{2}= \overline{P_{1}} \cap \overline{P_{2}} \cap \dots	\overline{P_{S-1}} \cap {P_{S}}\\
		   	& \dots\\
		   	& P'_{K}= P_{1} \cap P_{2} \cap \dots	P_{S-1} \cap P_{S}\\
    \end{aligned}
\end{equation}

where $P_{s},=\left\{e\in E,\: e\: fulfills\: P_{s}\right\}$ and $\overline{P_{s}}=\left\{e\in E, \:e \:does \:not \:fulfill \:  P_{s}\right\}$. 

Therefore, $\displaystyle \bigcap^{K}_{k=1} P'_{k}=\emptyset$ and $\displaystyle \bigcup^{K}_{k=1} P'_{k}=E$.\par

\begin{figure}[htb!]
	\begin{center}
		\includegraphics[width=6cm,keepaspectratio=true]{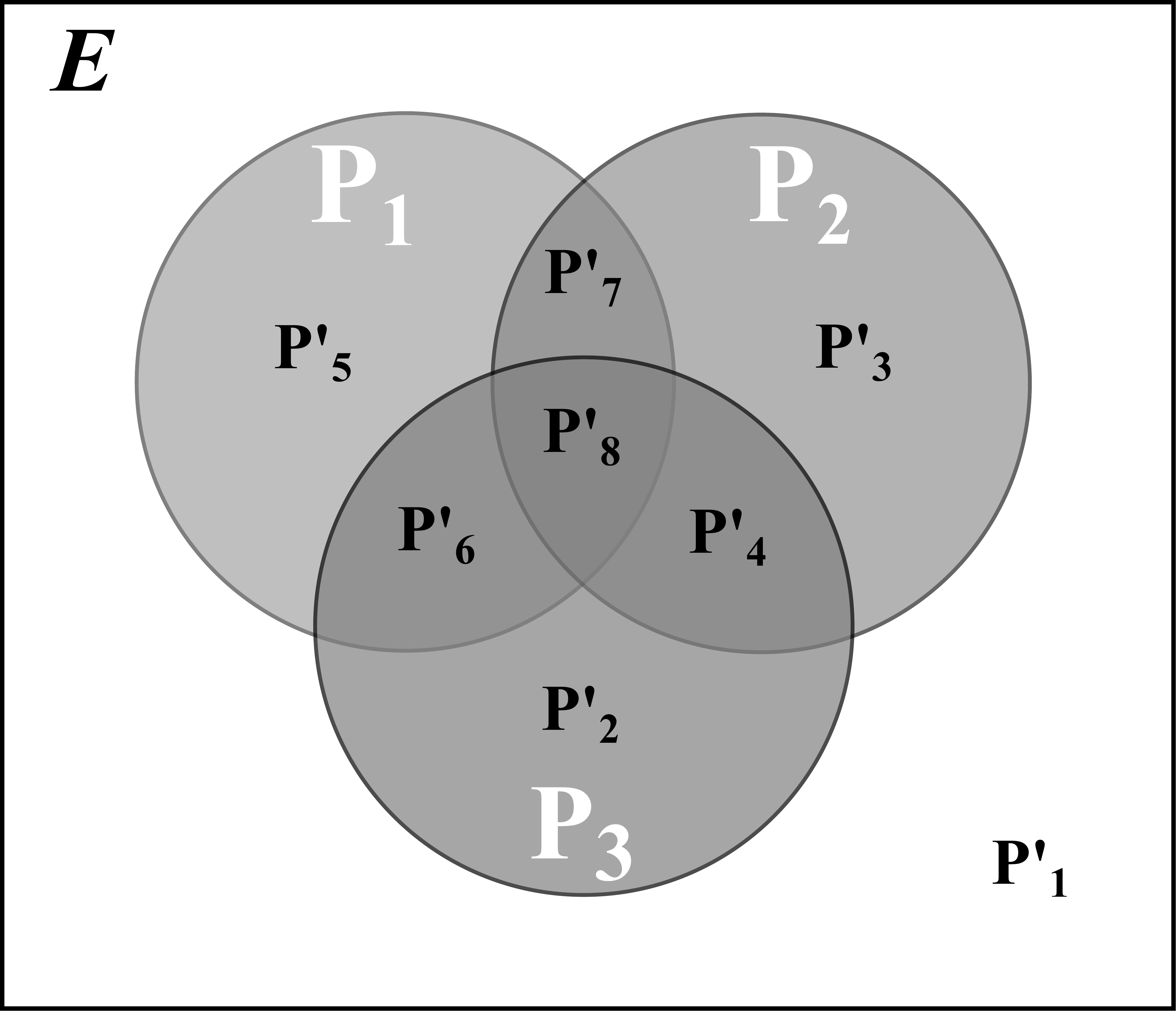}
	\end{center}
	\caption{Division of the referential universe for the case of $S=3$ properties and $8=2^3$ elements in the partition $P'_{k},\:k=1,\dots,8$. $P_{1}=P'_{5} \cup P'_{6} \cup P'_{7} \cup P'_{8}; P_{2}=P'_{3} \cup P'_{4} \cup P'_{7} \cup P'_{8}; P_{3}=P'_{2} \cup P'_{4} \cup P'_{6} \cup P'_{8}$}
	\label{fig:DivisionUniverseDisjointSets}
\end{figure}

Figure~\ref{fig:DivisionUniverseDisjointSets} shows and example for $S=3$ properties, where the $P'_{k}$, $k=1,\dots,8$ denote each one of the disjoint sets that are generated from the properties $P_1, P_2$ y $P_3$. So, for instance $P'_{8}=P_1 \cap P_2 \cap P_3$, $P'_{2}\cup P'_{6} = \overline {P_2} \cap P_3$ or $P_1$=$P'_{5} \cup P'_{6} \cup P'_{7} \cup P'_{8}$.\par

\subsubsection{Transformation into inequations}
\label{ssc:TransformationInequations}
The syllogistic problem is transformed into an equivalent optimization problem, disregarding the type of quantifiers involved; notwithstanding, we can deal at least with all types of quantifiers compatible with the linguistic TGQ (defined in Table~\ref{tab:FamiliesGeneralizedQuantifiers}).\par

In this section we describe the transformation with quantifiers defined as crisp intervals  ($Q=[a,b]$), since this is the basis for describing the reasoning process for fuzzy quantifiers (section~\ref{sc:ExamplesSyllogisms}). For the sake of simplicity, we assume that the $L_{i,j}s$ are atomic; that is, they are not boolean combinations of sets. This assumption does not limit the the generality of the solution that can be extended by simply and directly substituting the atomic parts by the boolean combination of sets. For instance, for $S=2$ we have the following combinations of disjoint sets:


\begin{equation}
	\begin{aligned}
	P'_{1}& = \overline{L_{1}} \cap \overline{L_{2}}\\
	P'_{2}& = L_{1} \cap \overline{L_{2}}\\
	P'_{3}& = L_{1} \cap L_{2}\\
	P'_{4}& = \overline{L_{1}} \cap L_{2}	
	\end{aligned}
\label{eq:ConjuntosDisjuntos}
\end{equation}

Where the usual notation $L_1,L_2$ for the terms in the propositions has been used instead of the one introduced in expression~\ref{eq:GeneralPatternSyllogism}.
Denoting $x_{k}=|P'_{k}|$, $k=1,\dots,K$ we have therefore,
\begin{equation}
	\begin{aligned}
		x_{1}&=\left| \overline{L_{1}} \cap \overline{L_{2}} \right|\\
		x_{2}&=\left| L_{1} \cap \overline{L_{2}} \right|\\
		x_{3}&=\left| L_{1} \cap L_{2} \right|\\
		x_{4}&=\left| \overline{L_{1}} \cap L_{2} \right|\\		
	\end{aligned}
\label{eq:CardinalityQuantifiers}
\end{equation}

Now, the evaluation of any quantified proposition is equivalent to solving an inequation where the previously defined cardinality values $x_{k}, k=1,\ldots,K$ are involved as variables. For instance, solving a binary proposition such as ``between 3 and 6 $Y_{1}$ are $Y_{2}$'', that involves the absolute binary quantifier $Q_{AB}=[3,6]$, is equivalent to inequations $x_3 \geq {4}$ and $x_{3} \leq 6$ (see (\ref{eq:ConjuntosDisjuntos})  and (\ref{eq:CardinalityQuantifiers})). Table~\ref{tab:FamiliesGeneralizedQuantifiersInequations} (right) summarizes the inequations corresponding to each quantifier.

\begin{table}[htb]
	\begin{center}
		\caption{Definitions of interval crisp Generalized Quantifiers $Q=[a,b]$}
		\vspace{2mm}
		\label{tab:FamiliesGeneralizedQuantifiersInequations}
			\begin{tabular}{|rl|c|} \hline
			\multicolumn{2}{|c|}{\textbf{Logical definition}} & \textbf{Equivalent inequation} \\\hline
			$Q_{LQ-all} (Y_{1}Y_{2})=$ & $\left\{\begin{array}{ll}
																			0:&Y_{1} \not\subseteq Y_{2}\\
																			1:&Y_{1} \subseteq Y_{2} \end{array} \right.$
																	& $x_{2}=0$ \\\hline
			$Q_{LQ-none} (Y_{1},Y_{2})=$ & $\left\{\begin{array}{ll}
																				0:&Y_{1} \cap Y_{2} \neq \emptyset \\
																				1:&Y_{1} \cap Y_{2}= \emptyset \end{array} \right.$
																	& $x_{3}=0$ \\\hline
			$Q_{LQ-some} (Y_{1},Y_{2})=$ & $\left\{\begin{array}{ll} 
																				0:&Y_{1} \cap Y_{2}=\emptyset \\
																				1:&Y_{1} \cap Y_{2}\neq \emptyset \end{array} \right.$
																	& $x_{3}>0$ \\\hline
			$Q_{LQ-not-all} (Y_{1},Y_{2})=$ & $\left\{\begin{array}{ll}
																									0:&Y_{1} \subseteq Y_{2}\\
																									1:&Y_{1} \not\subseteq Y_{2} \end{array} \right.$
																	& $x_{2}>0$ \\\hline
			$Q_{AB}(Y_{1},Y_{2})=$ & $\left\{\begin{array}{ll}
																	0:&|Y_{1} \cap Y_{2}| \notin [a,b]\\
																	1:&|Y_{1} \cap Y_{2}| \in [a,b] \end{array} \right.$
																	& $x_{3} \geq a ; x_{3} \leq b$\\\hline
			$Q_{PB}(Y_{1},Y_{2})=$ & $\left\{\begin{array}{ll}
																	0:&\frac{|Y_{1} \cap Y_{2}|}{|Y_{1}|} < a \vee \frac{|Y_{1} \cap Y_{2}|}{|Y_{1}|} > b\\
																	1:& |Y_{1}|=0\\
																	1:& \frac{|Y_{1} \cap Y_{2}|}{|Y_{1}|} \geq a \wedge \frac{|Y_{1} \cap Y_{2}|}{|Y_{1}|} 																				\leq b \end{array} \right.$
																	& $\begin{array}{l}
																				\frac{x_{3}}{x_{2}+x_{3}} \geq a;\\ 																																													\frac{x_{3}}{x_{2}+x_{3}} \leq b \end{array}$ \\\hline
			$Q_{EB}(Y_{1},Y_{2})=$ & $\left\{\begin{array}{ll}
																	0:& |Y_{1} \cap \overline{Y_{2}}| \notin [a,b]\\
																	1:& |Y_{1} \cap \overline{Y_{2}}| \in [a,b] \end{array} \right.$
																	& $x_{2} \geq a ; x_{2} \leq b$\\\hline																											
			$Q_{CB-ABS}(Y_{1},Y_{2})=$ & $\left\{\begin{array}{ll}
																			0:& |Y_{1}| - |Y_{2}| \notin [a,b]\\
																			1:& |Y_{1}| - |Y_{2}| \in [a,b] \end{array} \right.$
																	& $\begin{array}{l} 
																					(x_{3}+x_{2})-(x_{3}+x_{4}) \geq a;\\																																													(x_{3}+x_{2})-(x_{3}+x_{4}) \leq b \end{array}$\\\hline
			$Q_{CB-PROP}(Y_{1},Y_{2})=$ & $\left\{\begin{array}{ll}
																			0:& \frac{|Y_{1}|}{|Y_{2}|} \notin [a,b]\\
																			1:& \frac{|Y_{1}|}{|Y_{2}|} \in [a,b] \end{array} \right.$
																	& $\frac{\left(x_{3} + x_{2}\right)}{\left(x_{3} + x_{4}\right)} \geq a;
																		\frac{\left(x_{3} +	x_{2}\right)}{\left(x_{3} + x_{4}\right)} \leq b$\\\hline
			$Q_S(Y_{1},Y_{2})=$ & $\left\{\begin{array}{ll}
															0:& \frac{|Y_{1}\cap Y_{2}|}{|Y_{1}\cup Y_{2}|} < a, Y_{1}	\cup Y_{2} \neq \emptyset\\
															1:& \frac{|Y_{1}\cap Y_{2}|}{|Y_{1}\cup Y_{2}|} \geq a, Y_{1} \cup Y_{2} \neq \emptyset\\																			1:& Y_{1} \cup Y_{2} = \emptyset \end{array} \right.$
																	& $	\frac{x_{3}}{x_{2} + x_{3}+x_{4}} \geq a ; 
																			\frac{x_{3}}{x_{2} + x_{3}+x_{4}} \leq b$\\\hline
		\end{tabular}
	\end{center}
\end{table}

\subsubsection{Definition and resolution of the equivalent optimization problem}
\label{ssc:DefinitionResolutionOptimization}
Once defined the inequations corresponding to the statements involved in a syllogism, we are in conditions of approaching the reasoning problem as an equivalent mathematical optimization problem. The fundamental idea of this transformation is described in~\cite{Dubois1988,Dubois1990} for binary proportional quantifiers. However, our approach is more general since we also incorporate other type of quantifiers and also any boolean combination in the restriction and scope of the quantified statements.\par

In order to correctly apply the resolution method (SIMPLEX), it is necessary to add three additional constraints to the set of inequations obtained from the syllogistic argument. The first constraint guarantees that there are no sets with a negative number of elements;
			\begin{equation}
				x_{k} \geq 0, \forall k=1,\ldots,K=2^{S}	
			\label{eq:RestriccionesInecuaciones_1}
			\end{equation}

The other two constraints are only necessary if the quantifier of the conclusion is a proportional one. In this case, since the function to be optimized is a rational one, it should be guaranteed that;
\begin{itemize}
	\item there are no $0$ in the denominator of the involved fractions in order to avoid indefinition in the results; that is, $L_{n,1} \neq \emptyset$. So if we denote by $P'^{n,1}_{1} \cdots P'^{n,1}_{r}$ the disjoint parts of $L_{n,1}$, and by $x_{r}^{n,1}$ the cardinalities of the disjoint sets, the following must hold:
			\begin{equation}
				x_{1}^{n,1}+\cdots +x_{r}^{n,1} >0, \forall n=1,...,N	
			\label{eq:RestriccionesInecuaciones_2}
			\end{equation}
			
	\item the sum of the cardinalities must equal the cardinality of the referential universe:
			\begin{equation}
				\sum_{k=1}^K x_{k}									=		\left|E\right|\\
			\label{eq:RestriccionesInecuaciones_3a}
			\end{equation}
\end{itemize}

The conclusion of the syllogistic argument is the statement ``$Q_{C}$ $L_{C,1}$ are $L_{C,2}$'' (as indicated in \ref{eq:GeneralPatternSyllogism}). For the case of absolute quantifiers $Q_{C}=[a,b]$ the expressions that have to be optimized are the following ones:

\begin{eqnarray}
a_{m} = minimize & x_{m,n_{1}} + \ldots + x_{m,n_{I}}\\
b_{m} = maximize & x_{m,n_{1}} + \ldots + x_{m,n_{I}}
\label{eq:OptimizationFunction}
\end{eqnarray}
with $x_{m,n_{i}}\in\left\{ x_k, k=1,...,K\right\} \forall i=1,...,I$
\noindent
subject to the following restrictions: the premises of the syllogisms (from $PR1$ to $PRN$ as in explained in Sec.~\ref{ssc:TransformationInequations}) and the restriction stated in (\ref{eq:RestriccionesInecuaciones_1}). For the case of proportional quantifiers in the conclusion, we have:
\begin{eqnarray}
a_{m} = minimize & \frac{x_{m,n_{1}} + \ldots + x_{m,n_{I}}}{x_{m,d_{1}} + \ldots + x_{m,d_{J}}}\\
b_{m} = maximize & \frac{x_{m,d_{1}} + \ldots + x_{m,n_{I}}}{x_{m,d_{1}} + \ldots + x_{m,d_{J}}}
\label{eq:OptimizationFunctionProportional}
\end{eqnarray}

with $x_{m,n_{i}}, x_{m,d_{j}}\in\left\{ x_k, k=1,...,K\right\} \forall i=1,...,I, \forall i=j,...,J$, and subject to the following restrictions: premises of the syllogism and the three additional constraints (\ref{eq:RestriccionesInecuaciones_1}), (\ref{eq:RestriccionesInecuaciones_2}) and (\ref{eq:RestriccionesInecuaciones_3a}).\par

The optimization procedure depends on the types of quantifiers that are being used. In the following section we will pay attention to the following three cases: crisp interval quantifiers, fuzzy quantifiers approximated with two intervals and fuzzy quantifiers.\par

\section{Cases of different types of syllogisms with different definitions of quantifiers}
\label{sc:ExamplesSyllogisms}
We manage three different interpretations of the quantifiers: 
\begin{itemize}
	\item crisp interval quantifiers: a crisp quantifier $Q$ defined as an interval $[a,b]$. The case of precise quantifiers also can be managed as the particular case when $a=b$;
	\item fuzzy quantifiers approximated as pairs of intervals: a fuzzy quantifier $Q$ defined as a pair of intervals $\left\{KER_{Q},SUP_{Q}\right\}$, where $KER_{Q}=[b,c]$ represents the kernel and $SUP_{Q}=[a,d]$ the support of $Q$;
	\item fuzzy quantifiers $Q$ represented in the usual trapezoidal form with parameters $\left[a,b,c,d\right]$.
\end{itemize}

In the following sections, we describe the behaviour of our approach by using some examples of fuzzy syllogisms.

\subsection{Syllogisms using crisp interval quantifiers}
\label{ssc:SyllogismsIntervalQuantifiers}
This is the simplest approach and the basis for solving the other two. The optimization technique depends on the quantifier of the conclusion. Absolute, exception and absolute comparative cases can be calculated using the SIMPLEX optimization method, since the interval to be optimized depends on linear operations. In the case of proportional, comparative proportional and similarity quantifiers, \textit{linear fractional-programming} techniques must be used~\cite{Boyd2004}.\par

\subsubsection{Example 1: Dogs, cats and parrots}
\label{sssc:DogsCatsParrots}

As a first example, we consider the following problem\footnote{Problem taken from http://platea.pntic.mec.es/jescuder/mentales.htm}:

\begin{quote}
\textit{Dogs, cats and parrots}. How many animals do I have in my home, if all but two are dogs, all but two are cats and all but two are parrots? 
\end{quote} 

As we can see, a combination of exception crisp quantifiers are involved. In order to formalize the wording according to the typical form of a syllogism, the first step is to identify the number of involved terms. In this case, for simplicity and for coherence with Fig.~\ref{fig:DivisionUniverseDisjointSets}, let us consider \textit{$E=$animals in my home} and three terms $dogs=P_{1}$, $cats=P_{2}$ and $parrots=P_{3}$.\par

Table~\ref{tab:SyllogismExceptionLogicQuantifiers} shows a first formalization of the problem considering only the explicit information of the wording. The corresponding quantifiers in $PR1$, $PR2$ and $PR3$ are Binary of Exception ($Q_{EB}$) as defined in Table~\ref{tab:FamiliesGeneralizedQuantifiersInequations} (left) with $[a,b]=[2,2]$.\par

\begin{table}
	\centering
		\begin{tabular}{rl}
			$PR1:$& All animals but two are dogs\\ 
			$PR2:$& All animals but two are cats\\
			$PR3:$& All animals but two are parrots\\\hline
			$C:$  & There are $Q_C$ animals 
		\end{tabular}
	\caption{Syllogism with quantifiers of exception.}
	\label{tab:SyllogismExceptionLogicQuantifiers}
\end{table}

According to Table~\ref{tab:FamiliesGeneralizedQuantifiersInequations} (right), each one of the premises generates the following system of inequations:
\begin{equation}
	\begin{aligned}
		PR1 :& x_{1} + x_{2} + x_{3} + x_{4}  = 2;\\
		PR2 :& x_{1} + x_{2} + x_{5} + x_{6}  = 2;\\
		PR3 :& x_{1} + x_{3} + x_{5} + x_{7}  = 2;\\
	 	C :& x_{1} + x_{2} + x_{3}+x_{4}+x_{5}+x_{6}+x_{7}+x_{8} = a;\\
	\end{aligned}
\end{equation}

\noindent
where $Q_{C}=[a,b]$ is the quantifier in the conclusion. Applying the SIMPLEX method, we obtain, nevertheless, $Q_{C}=[0,inf)$; that is, an undefined result. The origin of this result is that the set of premises does not contain all the necessary information that human beings manage to solve this problem. It is necessary to incorporate additional premises with the implicit or contextual information. In this case, we had added five premises ($PR4-PR8$) thus producing the extended syllogism shown in Table~\ref{tab:SyllogismExceptionLogicQuantifiersAdditionalPremises}. 

\begin{table}
	\centering
		\begin{tabular}{rl}
			$PR1:$& All animals but two are dogs\\ 
			$PR2:$& All animals but two are cats\\
			$PR3:$& All animals but two are parrots\\
			$PR4:$& No dog, cat or parrot is not an animal\\
			$PR5:$& No animal is not a dog, a cat or a parrot\\
			$PR6:$& No dog is a cat or a parrot\\
			$PR7:$& No cat is a dog or a parrot\\
			$PR8:$& No parrot is a dog or a cat\\\hline
			$C:$  & There are $Q_C$ animals 
		\end{tabular}
	\caption{Syllogism with quantifiers of exception, logic and additional premises.}
	\label{tab:SyllogismExceptionLogicQuantifiersAdditionalPremises}
\end{table}

The corresponding quantifiers of the additional premises $PR4-PR8$ are logical quantifiers $Q_{4}=Q_{5}=Q_{6}=Q_{7}=Q_{8}=none$, labelled as $Q_{LQ-none}$ in Table~\ref{tab:FamiliesGeneralizedQuantifiersInequations} (left). According to Table~\ref{tab:FamiliesGeneralizedQuantifiersInequations} (right), the corresponding system of inequations is:
\begin{equation}
	\begin{aligned}
	PR4 :& x_{1} = 0 ;\\
	PR5 :& x_{1} = 0 ;\\
	PR6 :& x_{6} + x_{7} + x_{8} = 0;\\
	PR7 :& x_{4} + x_{7} + x_{8} = 0;\\
	PR8 :& x_{4} + x_{6} + x_{8} = 0;\\
	C :& x_{2} + x_{3}+x_{4}+x_{5}+x_{6}+x_{7}+x_{8} = a;\\
	\end{aligned}
\end{equation}

Applying the SIMPLEX method, we obtain $Q_{C}=[3,3]$; that is, ``\textit{There are $three$ animals}''; that is, I have a dog, a cat and a parrot in my home.\par

On the other hand, it is relevant to note that without the premises $PR4$ and $PR5$, we obtain $Q_{C}=[2,3]$; that is, without these constraints we are assuming the possibility of the existence of animals in my home that are not dogs, cats or parrot. Nevertheless, the wording of the problem has implicit that the only animals that can be at home are dogs, cats and parrots and for that reason they are included.\par

The example shows a relevant aspect to manage in problems expressed in natural language: contextual and implicit information must be incorporated to the syllogism. Those syllogistic patterns limited to predefined patterns or with a small number of premises cannot manage this kind of problems.\par

\subsubsection{Example 2: Students of sixth course}
\label{sssc:StudentsSixthCourse}
The following example is mainly focused on the use of interval quantifiers. The exercise is extracted from the Spanish Mathematical Olympiad 1969-70\footnote{\textsl{http://platea.pntic.mec.es/csanchez/olimp\_1963-2004/OME2004.pdf}; p. 29}:
\begin{quote}
In the tests of sixth course in a primary school, at least $70\%$ students passed the subject of Physics; at least $75\%$ students passed Mathematics; at least $90\%$ passed Philosophy and at least $85\%$ passed Foreign Language. How many students, at least, did pass these subjects?
\end{quote}

In this case, five terms must be considered in the universe $E$: \textit{students of sixth course$=P_{1}$}, \textit{students that passed Physics$=P_{2}$}, \textit{students that passed Mathematics$=P_{3}$}, \textit{students that passed Philosophy$=P_{4}$} and \textit{students that passed Foreign Language}$=P_{5}$. Table~\ref{tab:SyllogismTwo} shows a possible formalization, where the corresponding quantifiers are: $Q_{1}=[0.7,1]$ of $PR1$; $Q_{2}=[0.75,1]$ of $PR2$; $Q_{3}=[0.9,1]$ of $PR3$; $Q_{4}=[0.85,1]$ of $PR4$; and $Q_{C}=[a,b]$ the quantifier of the conclusion.\par

\begin{table}
	\centering
	\scalebox{0.95}{
		\begin{tabular}{rl}
			$PR1:$&At least $70\%$ of sixth course students passed Physics\\
			$PR2:$&At least $75\%$ of sixth course students passed Mathematics\\ 
			$PR3:$&At least $90\%$ of sixth course students passed Philosophy\\
			$PR4:$&At least $85\%$ of sixth course students passed Foreign Language\\\hline
			$C:$ &$Q_C$ \small sixth course students passed Physics, Mathematics, Philosophy and Foreign Language
		\end{tabular}
		}
	\caption{Syllogism five terms and five premises.}
	\label{tab:SyllogismTwo}
\end{table}

In this case, we avoid the details regarding the corresponding set of inequations of each premise. According to section~\ref{ssc:DivisionDisjointSets}, $2^{5}=32$ disjoint sets must be generated and the corresponding inequations. The quantifier of the conclusion is of type binary proportional and it must be calculated applying \textit{linear fractional programming} obtaining $Q_{C}=[0.2,1]$; that is ``At least $20\%$ of sixth course students passed Physics, Mathematics, Philosophy and Foreign Language'', which corresponds to the expected result.\par

\subsection{Syllogisms with fuzzy quantifiers approximated as pairs of intervals}
\label{ssc:FQuantifiersPairsIntervals}
In this method, each fuzzy quantifier $Q$ in the syllogism is defined as $Q=\left\{KER_{Q},SUP_{Q}\right\}$, where $KER_{Q}=[b,c]$ corresponds with the kernel and $SUP_{Q}=[a,d]$ corresponds with the support. In this model, the solution is obtained from two systems of inequations, the first of them taking $KER_{Q}$ as crisp interval definitions for all the statements in the syllogism and the second one taking $SUP_{Q}$. The corresponding solutions define $KER_{Q_C}$ and $SUP_{Q_C}$ for the quantifier $Q_C$ in the conclusion, respectively. We should note that this is an exact and very simple approach for cases where the quantifiers are trapezoids. For the cases of non-trapezoidal quantifiers this produces an approximate solution. In the cases where non-normalized quantifiers may appear (either in the definition of the premises or due to the non-existence of solutions for the inequations system) this approach cannot produce results and therefore solutions should be obtained by using the approach described in section~\ref{ssc:SyllogismsFuzzyQuantifiers}.\par  

For the example in Table~\ref{tab:SyllogismTwo} we have $Q_{1}=\left\lbrace [0.8,0.9],[0.7,1] \right\rbrace$; $Q_{2}=\left\lbrace [0.8,0.85],[0.75,0.9] \right\rbrace$; $Q_{3}=\left\lbrace [0.92,1],[0.9,1] \right\rbrace$; $Q_{4}=\left\lbrace [0.9,0.95],[0.85,1] \right\rbrace$; and $Q_{C}=\left\lbrace[b,c],[a,d]\right\rbrace$ the quantifier of the conclusion.\par



\begin{figure}
	\centering
		\includegraphics{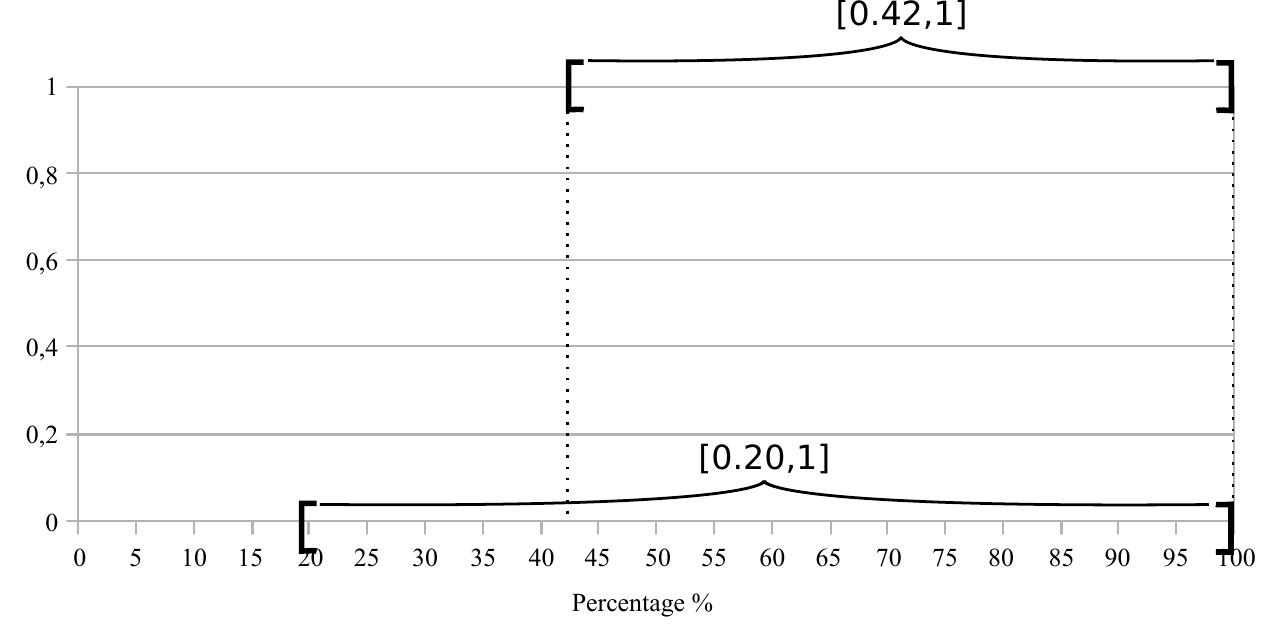}
	\caption{Graphical representation of the solution obtained for $Q_{C}$ for the example in Table~\ref{tab:SyllogismTwo} using the pairs of intervals approximation for the quantifiers in the syllogism.}
	\label{fig:FQs2Intervals}
\end{figure}

Applying the procedure described in section~\ref{ssc:SyllogismsIntervalQuantifiers} for each system, we obtain $KER_{Q_{C}}=[0.42,1]$ for the system defined by taking the \textit{Kernels} of all the quantifiers in the premises and $SUP_{Q_{C}}=[0.20,1]$ for the the system defined by taking the \textit{Supports} of all the quantifiers in the premises. The result is therefore $Q_{C}=\left\{[0.42,1],[0.2,1] \right\}$ which is a  result that entails the definition for $Q_C$=`at least 20\%' obtained in the previous section. Figure~\ref{fig:FQs2Intervals} shows the graphical representation of the quantifier $Q_{C}$, where the interval $[0.2,1]$ denotes the support of the fuzzy set and the interval $[0.42,1]$ denotes its kernel. \par

As we can see, using this approach we can only calculate the two represented intervals. Any other value between them must be interpolated. The main problem of this approach is for fuzzy quantifiers that are not represented as trapezoidal functions or those that are non-normalized. In these cases, we must use fuzzy quantifiers.

\subsection{Syllogisms with fuzzy quantifiers}
\label{ssc:SyllogismsFuzzyQuantifiers}

The use of fuzzy quantifiers supposes a generalization of the reasoning procedure described in the previous sections. Each fuzzy quantifier is managed through a number of $\alpha-$cuts, which are crisp intervals defined on the universe of discourse of the quantifier. Therefore, an inequations system that is similar to the ones in section \ref{ssc:TransformationInequations} is obtained for each $\alpha-$cut considered. The reasoning procedure consists of applying the method described in section~\ref{ssc:DefinitionResolutionOptimization} for each of these inequations systems. Each of the solutions obtained for all these systems define the corresponding $\alpha-$cut for the quantifier $Q_C$ in the conclusion. By using this general approach, it is possible to manage any generalized quantifier (as the ones shown in Table \ref{tab:FamiliesGeneralizedQuantifiers}) for three situations that are not considered in the previous models: 
\begin{itemize}
\item trapezoidal quantifiers where non-normalized results are obtained in any of the solutions, since for this case $KER_{Q_{C}}$ cannot be calculated and therefore the previous approach leads to indefinition.
\item quantifiers defined with non trapezoidal functions (in this case, this is an approximate solution where the level of approximation can be defined as wished) and 
\item typical linguistic fuzzy quantifiers like \textit{most, many, all but around three,\dots} 
\end{itemize}

\par

In order to better illustrate this approach we will consider the five types of examples described in the subsequent sections. 

\subsubsection{Example 1 (fuzzy extension): Students of sixth course}

\begin{figure}
	\centering
		\includegraphics{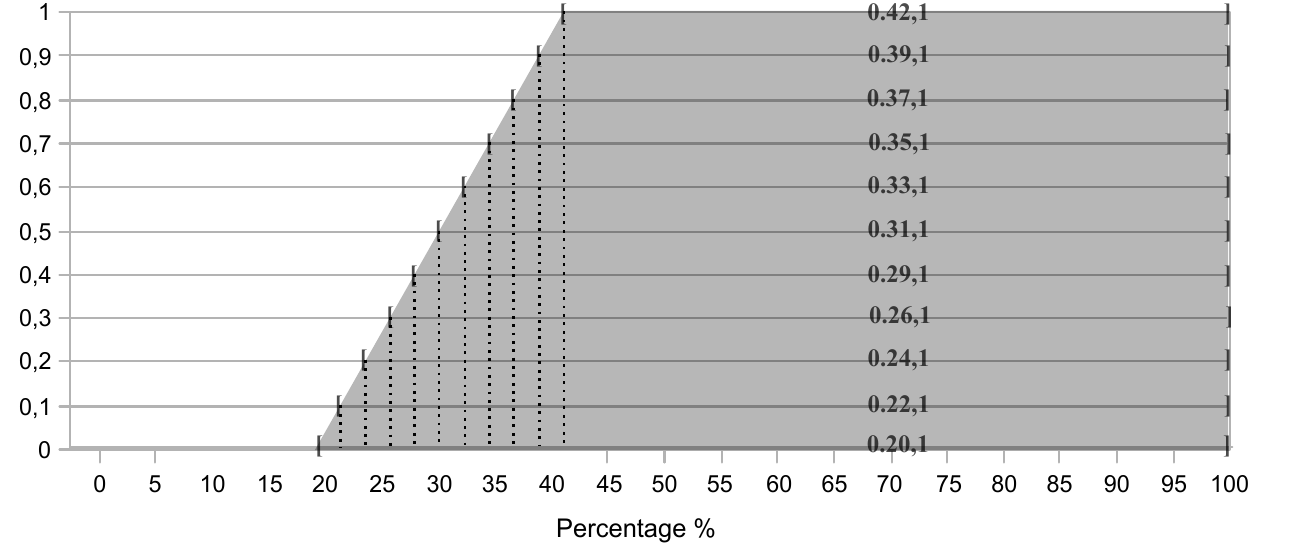}
	\caption{Graphical representation of $Q_{C}$ with 11 $\alpha-$cuts.}
	\label{fig:FQsAlfacuts}
\end{figure}

This first example is the one in Table~\ref{tab:SyllogismTwo} which is useful for showing the consistency of this approach with the one described in the previous section. We will consider fuzzy trapezoidal quantifiers that comprise the definitions for the quantifiers in Table~\ref{tab:SyllogismTwo}. Using the usual notation for trapezoidal fuzzy membership functions where $Q=[a,b,c,d]$ ($KER_{Q}=[b,c], SUP_{Q}=[a,d]$) we have: $Q_{1}=[0.7,0.8,0.9,1]$; $Q_{2}=[0.75,0.8,0.85,0.9]$; $Q_{3}=[0.9,0.92,1,1]$; $Q_{4}=[0.85,0.9,0.95,1]$; and $Q_{C}=[a,b,c,d]$ the quantifier of the conclusion. Applying the corresponding optimization method, we obtain $Q_{C}=[0.2,0.42,1,1]$, which is consistent with the results in the previous sections. Figure~\ref{fig:FQsAlfacuts} shows a fully graphical representation of $Q_{C}$ for the case considered involving eleven $\alpha$-cuts. As we can see, the eleven $\alpha$-cuts intervals allow us to better approximate a fuzzy definition for the conclusion quantifier $Q_C$. It is worth noting that the extreme cases, $\alpha-$cut$=0$ and $\alpha-$cut$=1$, correspond to the \textit{support} and \textit{kernel} cases in the approximation described in the previous section (see Figure~\ref{fig:FQs2Intervals}).\par

\subsubsection{Example 2 (non-normalized fuzzy extension): Students of sixth course}

In the second example, we illustrate a syllogism that produces a non-normalized fuzzy set as result. We consider again the example of Table~\ref{tab:SyllogismTwo} but adding the following premise:\par

\begin{center}
	$PR5:$ Between $40\%$ and $90\%$ sixth course students did not pass Physics or Mathematics or Philosophy or Foreign Language\par
\end{center}

\begin{figure}
	\centering
		\includegraphics{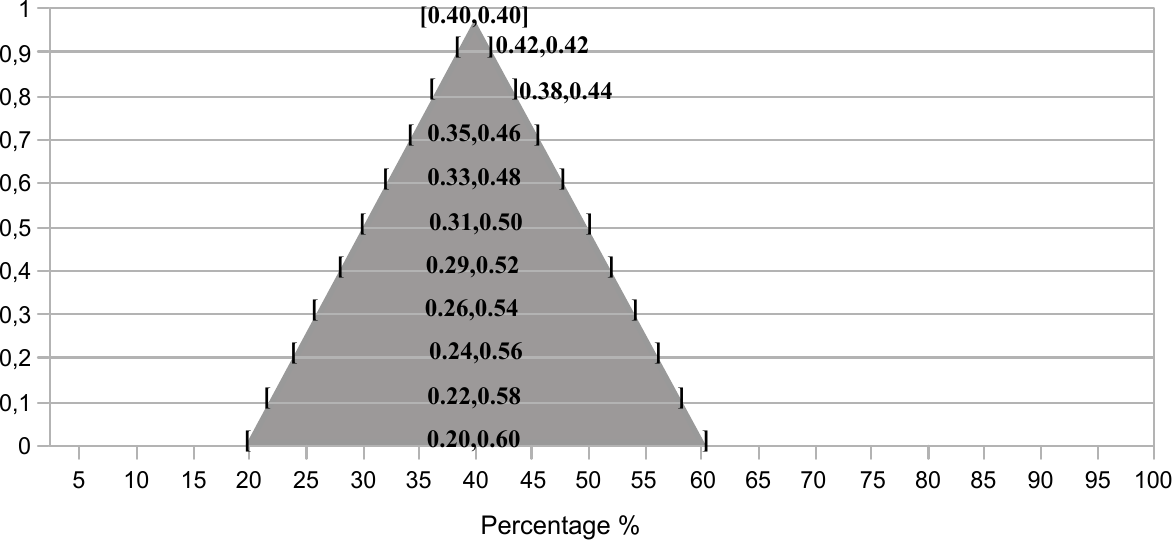}
		\caption{Graphical representation of the solution obtained for $Q_{C}$ for the example in Table~\ref{tab:SyllogismTwo} with the additional premise $PR4$.}
	\label{fig:FQsnon-normalized}
\end{figure}

Its definition is $Q_{5}=[0.4,0.6,0.8,0.9]$ and the corresponding system of inequations is added to the previous system.  Results are shown in Fig.~\ref{fig:FQsnon-normalized}. For $\alpha-$cuts higher than $0.95$ the system has no solution; that is, the fuzzy set that defines the quantifier of the conclusion is a non-normalized fuzzy set. Therefore, this syllogism cannot be solved managing fuzzy quantifiers approximated as pairs of intervals (section~\ref{ssc:FQuantifiersPairsIntervals}).

\subsubsection{Example 3 (non-trapezoidal result): wine warehouse}
In this case, we show a syllogism using the class of proportional quantifiers known as Regular Increasing Monotone (RIM) quantifiers~\cite{Yager1996}, that were proposed within the framework of quantifier guided aggregation of combination of criteria. This is a way of defining proportional quantifiers (such as \textit{few, many, most,\dots}) where the linguistic term is interpreted as a fuzzy subset $Q$ of the $[0, 1]$ interval. Its basic definition is shown in equation~\ref{eq:RIM}
\begin{equation}
	Q^{\alpha}(p)= p^{\alpha}; \alpha > 0
\label{eq:RIM}
\end{equation}
where $Q^\alpha$ denotes a linguistic quantifier and $p \in [0,1]$. For instance, for $Q^2$ (labelled as \textit{most}), $Q^2(0.95)=1$ means that saying 95\% completely fulfills the meaning conveyed by \textit{most} and $Q^2(0.6)=0.75$ means that saying 60\% fulfils with degree $0.75$ the meaning conveyed by \textit{most}. Other usual RIM quantifiers are $Q^{0.5}$, usually labelled~\cite{Malczewski2006} as ``a few'' and $Q^1$ (identity quantifier, also labelled as ``a half''). RIM quantifiers are also associated with the semantics of ``\textit{the greater the proportion of \dots the better}'' due to its increasing monotonic behaviour. Under this interpretation, the semantics should be labelled accordingly to the actual value of $\alpha$ as ``linear'' ($\alpha=1$), ``quadratical'' ($\alpha=2$), ``sub-linear'' ($\alpha<1$), ...\par

So, let us consider the following example using RIM quantifiers: in a wine warehouse that sells red, white wine and other products derived from grapes, we have the following sentences:

\begin{quote}
$Q^1$ bottles of red wine are sold in the United Kingdom. 
\\$Q^1$ bottles of red wine sold in the United Kingdom are bought by J. Moriarty.
\end{quote}

Both propositions involve the identity quantifier $Q^1$ ($\alpha=1$), that can either be interpreted as ``a half'' or as \textit{the greater\dots the better (linear, $\alpha=1$)}.


If we apply the Zadeh's~\cite{Zadeh1985} Intersection/product pattern, we can infer the following statement: ``$Q_{C}$ bottles of red wine are sold in United Kingdom and bought by J. Moriarty''. So, the corresponding syllogism is shown in Table~\ref{tab:SyllogismZadehIP}.\par
\begin{table}
	\centering
		\begin{tabular}{rl}
			$PR1:$& $Q^1$ bottles of red wine are sold in the United Kingdom \\
			$PR2:$& $Q^1$ bottles of red wine sold in the United Kingdom are bought by J. Moriarty \\\hline
			$C:$ &$Q_{C}$ bottles of red wine are sold in the United Kingdom and bought by J. Moriarty
		\end{tabular}
	\caption{Zadeh's Intersection/product syllogism.}
	\label{tab:SyllogismZadehIP}
\end{table}

\begin{figure}
	\centering
		\includegraphics{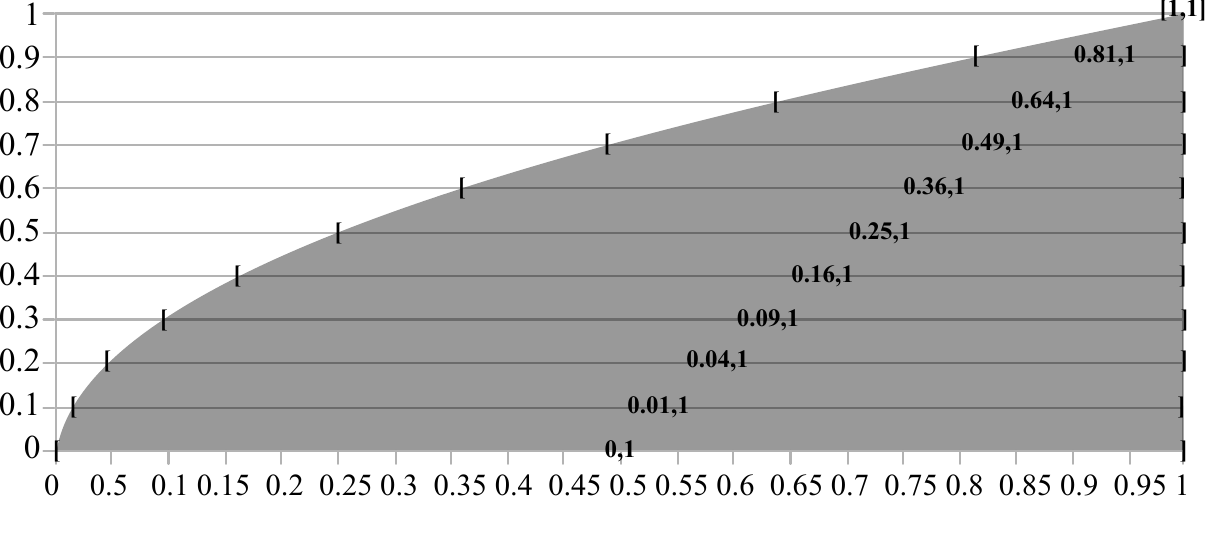}
	\caption{$Q_{c}$ of Example~\ref{tab:SyllogismZadehIP}}
	\label{fig:FQ_ZadehIP}
\end{figure}

As the result  of the corresponding optimization method, we obtain the quantifier $Q_C$ shown in Figure~\ref{fig:FQ_ZadehIP}, such that $Q_C=Q^{0.5}$. Its linguistic interpretations are ``\textit{a few}''~\cite{Malczewski2006} ``A few bottles of red wine sold in the United Kingdom are bought by J. Moriarty'' or ``The greater the proportion of bottles of red wine that are sold in the United Kingdom and bought by J. Moriarty the better'' (sublinear, $\alpha=0.5$). 

In case this example was solved using the $KER-SUP$ linear interpolation approach a much worse approximation (with bigger error) would be obtained for the inferred sublinear quantifier $Q_C=Q^{0.5}$  in Figure~\ref{fig:FQ_ZadehIP}.

\subsubsection{Example 4 (proportional fuzzy quantifiers): Account of a wine warehouse}
In this case, we show a syllogism involving fuzzy quantifiers of the type \textit{many, most,\dots}. Let us consider the account of the wine warehouse, that is not so good, and they only manage the following data:
\begin{quote}
Many sales are from red wine. A few sales are from white wine.
\end{quote}

Taking both statements as premises in the universe of \textit{products sold by wine warehouse=$E$}, we have a syllogism with three terms: \textit{sales=$P_{1}$}, \textit{red wine=$P_{2}$} and \textit{white wine=$P_{3}$}. With these two premises we can use, for example, consequent disjunction pattern of Zadeh~\cite{Zadeh1985}, which conclusion has the form ``$Q_{C}$ $P_{1}$ are $P_{2}$ or $P_{3}$'' being $P_{1}$ the subject of the premises and $P_{2}$ and $P_{3}$ the corresponding predicates. However, in this example, we change the conclusion by the following one ``$Q_{C}$ sales are not from red wine or white wine'', because this allow them to know how many sales are from other products. Table~\ref{tab:SyllogismZadehCD} shows the complete syllogism.\par

\begin{table}
	\centering
		\begin{tabular}{rl}
			$PR1:$& Many sales are from red wine\\
			$PR2:$& Few sales are from white wine\\ \hline
			$C:$ &$Q_{C}$ sales are not from red wine or white wine
		\end{tabular}
	\caption{Zadeh's Consequent Disjunction syllogism with different conclusion.}
	\label{tab:SyllogismZadehCD}
\end{table}

\begin{figure}
	\centering
		\includegraphics{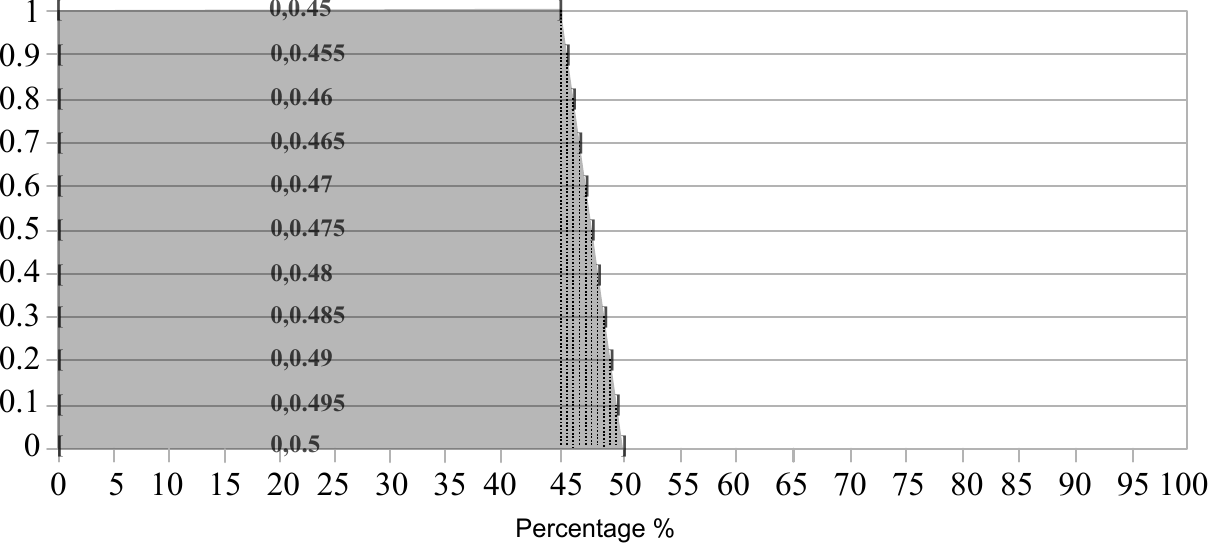}
		\caption{Graphical representation of the solution obtained for $Q_{C}$ for the example in Table~\ref{tab:SyllogismZadehCD}.}
	\label{fig:FQsalpha-cutsDecreasing}
\end{figure}

The first step to perform in the inference process is to assign the corresponding trapezoidal function to each one of the linguistic quantifiers of the premises. Taking as inspiration~\cite{Solt2011}, we assign the following trapezoids: $Many=[0.5,0.55,0.65,0.7]$ and $Few=[0.1,0.15,0.2,0.25]$. Next, we apply the reasoning procedure obtaining the following result: $Q_{C}=[0,0,0.45,0.5]$; that is, more or less, ``At most half sales are not from red wine or white wine''. In Fig.~\ref{fig:FQsalpha-cutsDecreasing} we show the corresponding fuzzy set. It is relevant to note that the obtained $Q_{C}$ of this example is a decreasing quantifier, a type of quantifier that some classical models of fuzzy syllogism cannot manage~\cite{Zadeh1985, Pereira2012}.\par

Example shown in Table~\ref{tab:SyllogismZadehCD} shares the same aspect of P. Peterson's Intermediate Syllogisms~\cite{Peterson2000}, but with a different conclusion because Peterson's schemas only include simple conclusions, without involving logical operations like disjunction. Table~\ref{tab:BKOSyllogism} shows the syllogism of Table~\ref{tab:SyllogismZadehCD}, but with different terms for a more clear example (\textit{people}, \textit{white hats} and \textit{red ties}) and according to BKO (B means that $PR1$ involves the quantifier ``few''; K that $PR2$ involves the quantifier ``many'', and O that $C$ involves the quantifier ``some\dots not'' ) Peterson's schema of Figure III. Using the same previous definitions for the \textit{Few} and \textit{Many} quantifiers, we obtain the following result: $Q_{C}=[0,0,0.98,0.98]$, that is consistent with the usual definition for the quantifier \textit{some\dots not}$=[0, 0, 1-\epsilon, 1-\epsilon]$.

\begin{table}
	\centering
		\begin{tabular}{rl}
			$PR1:$& Few people are wearing white hats\\
			$PR2:$& Many people are wearing red ties\\ \hline
			$C:$ &$Q_{C}$ people that are wearing red tie do not wear white hat
		\end{tabular}
	\caption{Peterson's BKO Syllogism.}
	\label{tab:BKOSyllogism}
\end{table}

\subsubsection{Example 5 (Exception fuzzy quantifiers): wine warehouse}
The last example shows a syllogism combining fuzzy quantifiers that have not been previously considered in the literature like exception quantifiers (i.e., \textit{all but around four,\dots}) combined with absolute ones (i.e., \textit{around four,\dots}):
\begin{quote}
All but around fifteen boxes of wine are for J. Moriarty. Around four boxes of the boxes that are not for J. Moriarty are for J. Watson.
\end{quote}

We can identify the following three terms in the statements: \textit{boxes of wine$=P_{1}$}, \textit{boxes for J. Moriarty$=P_{2}$} and \textit{boxes for J. Watson$=P_{3}$}. Given this information, we can infer how many boxes are not for J. Moriarty neither for J. Watson; that is, ``$Q_{C}$ boxes are not for J. Moriarty neither for J. Watson''. In Table~\ref{tab:SyllogismNonTypicalQ} is afforded the complete argument.\par
\begin{table}
	\centering
		\begin{tabular}{rl}
			$PR1:$& All but around fifteen boxes of wine are for J. Moriarty\\
			$PR2:$& Around four boxes of the boxes that are not for J. Moriarty are for J. Watson\\ \hline
			$C:$ &$Q_{C}$ boxes are not for J. Moriarty neither for J. Watson
		\end{tabular}
	\caption{Zadeh's Intersection/product syllogism.}
	\label{tab:SyllogismNonTypicalQ}
\end{table}

The quantifier of $PR1$ is a fuzzy quantifier of exception, because it is indicating that there is \textit{around fifteen} boxes of wine that are not for J. Moriarty. For instance, the associated trapezoidal function to this number can be \textit{around 15}$=[13,14,16,17]$. The quantifier of $PR2$ is another fuzzy quantifier, but in this case an absolute one. This kind of quantifiers is analyzed in the literature from the very beginning (Zadeh's distinction between absolute/proportional fuzzy quantifiers \cite{Zadeh1983}), but it has not been considered from the point of view of its use in reasoning. In this case, we assign \textit{around four}$=[3,4,4,5]$.\par

After applying the reasoning process, we obtain the absolute quantifier $Q_{C}=[8,10,12,14]$ for the conclusion, with the usual associated meaning of \textit{around 11} (since value $11$ is in the middle of the trapezoid's kernel). Therefore, in this example the conclusion states that ``Around eleven boxes are not for J. Moriarty neither for J. Watson''.\par

On the other hand, it is worth noting that this model is consistent with a fuzzy arithmetic approach to this problem; since from $PR1$ we have that ``around 15'' boxes are not for J. Moriarty, and from $PR2$ we have that ``around four'' of these are not for J. Watson; therefore, ``around 15'' $\ominus$ ``around 4'' $=$ ``around 11'', which is the number of boxes that are not for J. Moriarty neither for J. Watson.

\section{Conclusions}
\label{sc:Conclusions}
We have formulated a general approach to syllogistic reasoning that overcomes the limitation in the number of premises and is compatible with the TGQ, which allows us to manage new types of quantifiers not considered in the literature until now. Furthermore, we show a way for transforming the reasoning problem with binary quantified statements in a mathematical optimization problem for obtaining the conclusions of the syllogism. The approach also can manage different definitions of quantifiers, both crisp and fuzzy. This model is currently implemented as a software library that is able to manage all the types of quantifiers described in the paper.\par


As a future work, we aim to extend our method for managing different types of quantifiers within the same syllogism and to include new type of quantifiers of arities greater than two (ternary and quaternary ones). Other relevant topic that can be considered is the managing and formalization of contextual or implicit information that is relevant for the syllogistic reasoning.
 Finally, developing a symbolic approach to syllogism generating a table of syllogisms using a determined set of quantifiers (with a definition supported on their use in natural language) is another aim.

\section*{Acknowledgments}
This work was supported by the Spanish Ministry of Science and Innovation (grant TIN2008-00040), the Spanish Ministry for Economy and Innovation and the European Regional Development Fund (ERDF/FEDER) (grant TIN2011-29827-C02-02) and the Spanish Ministry for Education (FPU Fellowship Program). Authors also wish to thank A. Ramos and P. Montoto for the development of the software tool to test our model.

\bibliographystyle{elsarticle-num}
\bibliography{bibtese}

\begin{thebibliography}{10}
\expandafter\ifx\csname url\endcsname\relax
  \def\url#1{\texttt{#1}}\fi
\expandafter\ifx\csname urlprefix\endcsname\relax\def\urlprefix{URL }\fi
\expandafter\ifx\csname href\endcsname\relax
  \def\href#1#2{#2} \def\path#1{#1}\fi

\bibitem{Kneale1968}
W.~Kneale, M.~Kneale, The Development of Logic, Clarendon Press, Oxford, 1968.

\bibitem{Kumova2010}
B.~I. Kumova, H.~\c{C}akir,
  \href{http://dl.acm.org/citation.cfm?id=1927099.1927141}{The fuzzy
  syllogistic system}, in: Proceedings of the 9th Mexican international
  conference on Artificial intelligence conference on Advances in soft
  computing: Part II, MICAI'10, Springer-Verlag, Berlin, Heidelberg, 2010, pp.
  418--427.
\newline\urlprefix\url{http://dl.acm.org/citation.cfm?id=1927099.1927141}

\bibitem{Liu1998}
Y.~Liu, E.~E. Kerre, An overview of fuzzy quantifiers (ii). reasoning and
  applications, Fuzzy Sets and Systems 95 (1998) 135--146.

\bibitem{Yager1986}
R.~Yager, Reasoning with fuzzy quantified statements: {P}art {II}, Kybernetes
  15 (1986) 111--120.

\bibitem{Diaz2003}
F.~D\'{i}az-Hermida, A.~J. Bugar\'{i}n, S.~Barro, Definition and
  {C}lassification of {S}emi-fuzzy {Q}uantifiers for the {E}valuation of
  {F}uzzy {Q}uantified {S}entences, International Journal of Approximate
  Reasoning 34~(1) (2003) 49--88.

\bibitem{Glockner2006}
I.~Gl\"ockner, Fuzzy Quantifiers: A Computational Theory, Springer, Berlin,
  2006.

\bibitem{Aristotle1949}
Aristotle, Prior and posterior analytics, Clarendom Press, Oxford, 1949.

\bibitem{Peterson2000}
P.~L. Peterson, Intermediate {Q}uantities. {L}ogic, linguistics, and
  {A}ristotelian semantics, Ashgate, Alsershot. England, 2000.

\bibitem{Zadeh1985}
L.~A. Zadeh, Syllogistic reasoning in fuzzy logic and its applications to
  usuality and reasoning with dispositions, IEEE Transactions On Systems, Man
  and Cybernetics 15~(6) (1985) 754--765.

\bibitem{Dubois1988}
D.~Dubois, H.~Prade, On fuzzy syllogisms, Computational Intelligence 4~(2)
  (1988) 171--179.

\bibitem{Dubois1990}
D.~Dubois, H.~Prade, J.-M. Toucas, Intelligent Systems. State of the Art and
  Future Directions, Ellis Horwood, Great Britain, 1990, Ch. Inference with
  imprecise numerical quantifiers, pp. 52--72.

\bibitem{Dubois1993}
D.~Dubois, L.~Godo, R.~L\'opez~de M\'antaras, H.~Prade, Qualitative {R}easoning
  with {I}mprecise {P}robabilities, Journal of Intelligent Information Systems
  2 (1993) 319--363.

\bibitem{Novak2008}
V.~Novák, A formal theory of intermediate quantifiers, Fuzzy Sets and Systems
  159 (2008) 1229--1246.

\bibitem{Schwartz1997}
D.~G. Schwartz, Dynamic reasoning with qualified syllogisms, Artifical
  Intelligence 93 (1997) 103--167.

\bibitem{Pereira2012}
M.~Pereira-Fariña, J.~C. Vidal-Aguiar, P.~Montoto, F.~Díaz-Hermida, A.~Bugarín,
  Razonamiento silogístico aproximado con cuantificadores generalizados, in:
  Actas del XVI Congreso Español de Tecnologías y Lógica Fuzzy (ESTYLF2012),
  2012, pp. 701--706.

\bibitem{Sommers1982}
F.~Sommers, The {L}ogic of {N}atural {L}anguage, Clarendon Press, Oxford, 1982.

\bibitem{Spies1989}
M.~Spies, Syllogistic Inference under Uncertainty, Psychologie Verlags Union,
  1989.

\bibitem{Schwartz1996}
D.~G. Schwartz, On the semantics for qualified syllogisms, in: Proceedings of
  the Fifth IEEE International Conference on Fuzzy Systems : FUZZ-IEEE'96 :
  September 8-11, 1996, IEEE, New York, 1996.

\bibitem{Murinova2012}
P.~Murinová, V.~Novák, A formal theory of generalized intermediate syllogisms,
  Fuzzy Sets and Systems 186 (2012) 47--80.

\bibitem{Zadeh1983}
L.~A. Zadeh, A {C}omputational {A}pproach to {F}uzzy {Q}uantifiers in {N}atural
  {L}anguage, Computer and Mathematics with Applications 8 (1983) 149--184.

\bibitem{Zadeh1987}
L.~A. Zadeh, A theory of approximate reasoning, in: R.~R. Yager,
  S.~Ovchinnikov, R.~M. Tong, H.~T. Nguyen (Eds.), Fuzzy Sets and Applications:
  Selected Papers by L. A. Zadeh, John Wiley \& Sons, New York, 1987.

\bibitem{Pereira2010}
M.~Pereira-Fariña, F.~Díaz-Hermida, A.~Bugarín, An analysis of reasoning with
  quantifiers within the {A}ristotelian syllogistic framework, in: Proc. 2010
  {I}nternational {F}uzzy {C}onference on {F}uzzy {S}ystems ({FUZZ}-{IEEE}
  2010), IEEE, July, 18-23, 2010 - CCIB, Barcelona, Spain, 2010, pp. 635--642.

\bibitem{Pereira2012b}
M.~Pereira-Fariña, F.~Díaz-Hermida, A.~Bugarín,
  \href{http://www.sciencedirect.com/science/article/pii/S0165011412001418?v=s5}{On
  the analysis of set-based fuzzy quantified reasoning using classical
  syllogistics}, Fuzzy Sets and Systems~(0) (2012) --.
\newblock \href {http://dx.doi.org/10.1016/j.fss.2012.03.015}
  {\path{doi:10.1016/j.fss.2012.03.015}}.
\newline\urlprefix\url{http://www.sciencedirect.com/science/article/pii/S0165011412001418?v=s5}

\bibitem{Barwise1981}
J.~Barwise, R.~Cooper, Generalized quantifiers and natural language,
  Linguistics and Philosophy 4 (1981) 159--219.

\bibitem{Peterson1995}
P.~L. Peterson, Distribution and proportion, Journal of Philosophical Logic
  24~(2) (1995) 193--225.

\bibitem{Boyd2004}
S.~Boyd, L.~Vandenberghe, Convex Optimization, Cambridge University Press, The
  Edinburgh Building, Cambridge, UK, 2004.

\bibitem{Yager1996}
R.~R. Yager, Quantifier guided aggregation using owa operators, International
  Journal of Intelligent Systems 11~(1) (1996) 49--73.

\bibitem{Malczewski2006}
J.~Malczewski, Ordered weighted averaging with fuzzy quantifiers: Gis-based
  multicriteria evaluation for land-use suitability analysis, International
  Journal of Applied Earth Observation and Geoinformation 8 (2006) 270--277.

\bibitem{Solt2011}
S.~Solt, Understanding Vagueness. Logical, Philosophical and Linguistic
  Perspectives, Collegue Publications, 2011, Ch. Vagueness in quantity: Two
  case studies from a linguistic perspective, pp. 157--174.

\end{thebibliography}

\end{document}